# Can gamification reduce the burden of self-reporting in mHealth applications? A feasibility study using machine learning from smartwatch data to estimate cognitive load.

Michal K. Grzeszczyk[1,4], M.Sc.; Paulina Adamczyk[1,3], B.Sc.; Sylwia Marek[1,3], B.Sc.; Ryszard Pręcikowski[1,3], B.Sc.; Maciej Kuś[1,3], B.Sc.; M. Patrycja Lelujko[1], B.Sc.; Rosmary Blanco[1], M.Sc.; Tomasz Trzciński[4,5,6], D.Sc.; Arkadiusz Sitek[7], PhD; Maciej Malawski[1,3], D.Sc.; Aneta Lisowska[1,2], EngD

[1]Sano Centre for Computational Medicine, Cracow, Poland; [2]Poznań University of Technology, Poznań, Poland; [3]AGH University of Science and Technology, Cracow, Poland; [4]Warsaw University of Technology, Warsaw, Poland; [5]IDEAS NCBR, Warsaw, Poland; [6]Tooploox, Wroclaw, Poland; [7]Massachusetts General Hospital, Harvard Medical School, Boston, MA, USA

**Abstract**
*The effectiveness of digital treatments can be measured by requiring patients to self-report their state through applications, however, it can be overwhelming and causes disengagement. We conduct a study to explore the impact of gamification on self-reporting. Our approach involves the creation of a system to assess cognitive load (CL) through the analysis of photoplethysmography (PPG) signals. The data from 11 participants is utilized to train a machine learning model to detect CL. Subsequently, we create two versions of surveys: a gamified and a traditional one. We estimate the CL experienced by other participants (13) while completing surveys. We find that CL detector performance can be enhanced via pre-training on stress detection tasks. For 10 out of 13 participants, a personalized CL detector can achieve an F1 score above 0.7. We find no difference between the gamified and non-gamified surveys in terms of CL but participants prefer the gamified version.*

**Introduction**
Digital health interventions (DHI) support patient health monitoring and treatment provision outside of the traditional clinical setting. To assess patient health status and track the effectiveness of remote treatments patients are requested to regularly fill surveys which consist of numerous questions relating to their physical and mental health. However, responding to questionnaires is laborious and might discourage the consistent use of mobile health (mHealth) applications. Recently Oakley-Girvan *et al.*[13] performed scoping review of mobile interventions and identified application features that impact patient engagement. A high number of surveys were among the features decreasing engagement with the mHealth apps. This is a problem when health intervention outcome is associated with effective engagement with the application[4].

To encourage long-term engagement app developers use gamification[9]. The use of game elements such as progress tracking and rewards has shown to motivate users to be physically active[19] or maintain a healthy diet[3]. However, the introduction of game elements in mHealth surveys context has been less commonly studied. In this work, we investigate the impact of the inclusion of simple game elements in mobile surveys on completion time and the cognitive burden of self-reporting. We hypothesize that gamification can ease the process of survey completion and ultimately facilitate long-term well-being tracking. To test this hypothesis we train a machine learning tool for cognitive load detection. Then, we develop gamified and not-gamified versions of a mobile survey application and conduct a study with human subjects to assess the effort required to complete these surveys using the developed tool.

**Related work**
*Cognitive Load Detection.* To obtain a good and unobtrusive cognitive load detector, which could facilitate both development of timely context-aware reminders and the evaluation of application use effort in the wild[8], researchers considered the utilization of wearable devices and machine learning. Morkova *et al.*[12] used a Shimmer3 GSR+ device attached to the participant's two fingers and an ear lobe to capture galvanic skin responses (GSR), electrocardiograms (ECG) and PPG signal of individuals during cognitively demanding tasks and neutral condition (CLAS dataset). The authors trained a person-specific Support Vector Machine (SVM) model on features extracted from GSR either paired with features extracted from ECG or PPG recordings to predict subjects' concentration. They reported average classification performance of 78% (ECG+GSR) and 74% (PPG+GSR) and commented that these performance differences were not significant and do not justify the use of ECG. Consumer-grade wearables, employed in just-in-time interventions, due to high cost and low practicality rarely are equipped with ECG and GSR sensors commonly available in devices used in experimental conditions[15]. Therefore, we focus on developing a cognitive load detector that utilizes PPG signal only, that can be captured by smartwatches.

Lisowska *et al.*[11] use a PPG signal from CLAS dataset[12] to detect cognitive load. The authors compared the performance of SVM with a different number of features vs. simple 1D Convolutional Neural Network (1D CNN). The latter performed better, however, the achieved average cognitive load classification accuracy reached only 60%. The authors tried to improve cognitive load detection through semi-supervised model training leveraging additional unlabeled signal, however, no notable performance gains were reached. We base our solution on this approach and first investigate the limitation of the simple 1D CNN and later suggest a potential change to the training approach. We consider between-task transfer learning. Our investigation is based on the following observations:

1. Even though deep learning approaches have been shown to outperform traditional methods utilizing hand-crafted features such as SVMs in a range of applications[14], including cognitive load detection from PPG signal[8], CNN-based models are known to be data hungry and in low data regimes, they tend to be pre-trained to offer better performance[20].
2. Cognitive load annotations are not easy to gather and there are few datasets utilizing physiological signals from wrist-worn devices designed to capture that state. On the other hand, stress classification is a commonly tackled problem and high-quality publicly available physiological signal datasets, such as WESAD[16] (Wearable Stress and Affect Detection), provide signal paired with labels convenient for ML training.

***Well-being Survey Gamification.*** Gamification refers to the use of game elements such as progress tracking, avatars, points, levels and challenges in a non-game context. Carlier *et al.*[2] conducted a comparative study of user experience with gamified and non-gamified surveys and reported that they found no difference between conditions in terms of data quality gathered but respondents of gamified surveys perceived the time taken to respond to questions as shorter than respondents of non-gamified surveys. We consider real time taken to answer the question and time spent in high cognitive load.

In terms of game elements, Carlier *et al.* used progress tracking in gamified version and the remaining game elements were personalised to each individual. In our previous work, we investigated the personalisation of gamification design of mHealth surveys and found that the selected preferred game elements in that context do not vary strongly between individuals[1]. Therefore in this work, we use a uniform version of gamified survey app across all participants selecting game elements associated with the *Player* archetype such as progress tracking, points, rewards and levels.

**Methods**

To compare cognitive effort exert by gamified and non-gamified mobile surveys we: collected two datasets that will be used for the development of a cognitive load detector and evaluation of gamified survey app. The experimental flow is presented in Figure 1. In this section, we describe the development of cognitive load detector utilising raw PPG signal. We selected gamification features and implemented gamified and non-gamified mobile survey applications. Lastly, we applied the cognitive load detector to PPG signal captured during responding to gamified and non-gamified mobile surveys.

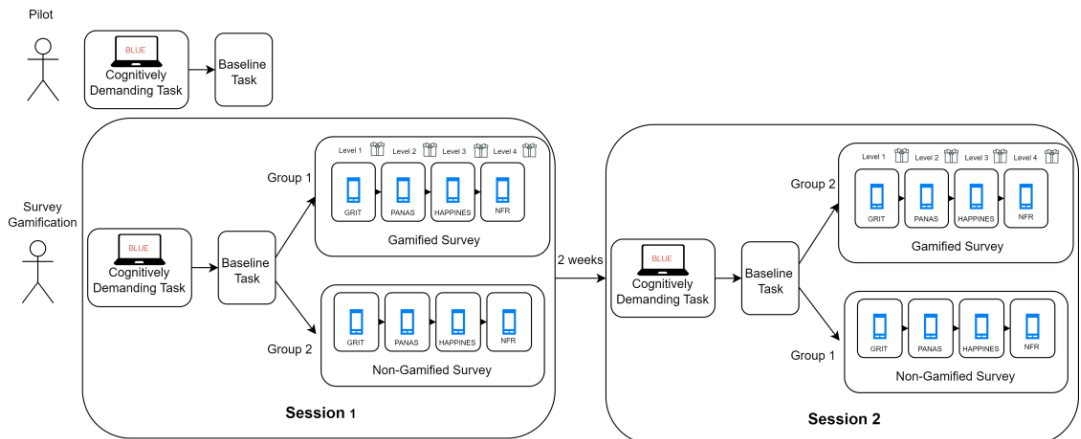

**Figure 1.** The process of Pilot and Survey Gamification datasets acquisition.

***Data Collection.*** The study was part of a larger well-being at-work study and was advertised through internal company email and Slack channels. We collected data from 24 healthy participants. Data from 11 participants were used for cognitive load model development and the remaining data have been used for evaluation of survey gamification. Participants performed all tasks in the seated position. The cognitively demanding task was conducted on the PC, while surveys were being completed on the provided smartphone.

***Pilot Dataset.*** To develop a cognitive load detector we gathered a pilot dataset capturing the physiological signal from 11 volunteers (5 females, 6 males aged 21-45) in two experimental conditions: cognitive load and baseline. In cognitive load condition, participants performed a Stroop test[8,] a cognitively demanding task in which participants are asked to promptly indicate colour of the text, but the text states the name of the colour that does not match the colour of the text, e.g. the word "red" printed in green instead of red ink. The Stroop task had 120 trials and we used PsyToolkit[17,18] implementation of the test. In the baseline neutral condition, participants were asked to sit and not perform any task. For the initial participants, the baseline condition lasted just a minute but later it has been adjusted to 3 minutes to match the average time needed to complete the Stroop test. In both conditions, participants wore Empatica4 device, on the wrist of their non-dominant hand, which recorded Blood Volume Pulse (BVP) (64 Hz), Electrodermal Activity (4 Hz), Temperature (4Hz), and Acceleration (32 Hz). For the purpose of further experiments, we used only BVP which can be captured by consumer-grade smartwatches.

***Survey Gamification Dataset***. To evaluate the impact of gamification on participants' cognitive load, we captured signal from 14 volunteers who first performed the same two tasks as in the pilot study and then completed either gamified or non-gamified survey. Each participant was planned to perform the experiment twice with a two weeks break between experimental sessions as a part of the work well-being study. To control for task familiarity, we divided the participants into two groups: the first group filled the gamified surveys in the first session and the non-gamified surveys in the second session, the second group had the opposite order. The participants were not informed that there will be a difference in the mobile survey application between sessions.

To facilitate the replication of our findings the pilot and survey gamification dataset has been made publicly available[6] on PhysioNet[5]. The study was approved by the AGH University of Science and Technology Ethics Committee. Note that one participant did not agree to make their data publicly available and two participants did not complete the second session. Therefore, in this work, we report only on results from 11 volunteers from the survey gamification experiment who participated in both conditions gamified and not (7 females and 4 males, age 26-55). There are 5 volunteers who filled gamified surveys in the first session and non-gamified in the second session and 6 volunteers who started from a non-gamified survey.

***Cognitive Load Detector***. Our model is a shallow 1D CNN, used previously for PPG signal classification in cognitive load detection[11] and stress classification tasks[7]. The model consists of 2 convolutional layers with 16 and 8 filters respectively and a kernel size of 3, followed by a max pooling layer and fully connected layer with 30 nodes and an output layer of size 2, corresponding to two target prediction classes: cognitive load or neutral. The model was implemented in Keras with the Tensorflow backend. We have not performed extensive parameter search concerning the number of layers nodes and kernel sizes but kept the parameters as reported in previous works[8,10].

***Training and Validation Protocol.*** We experimented with two training protocols: 1) Vanilla- training only on captured pilot cognitive load dataset and 2) WESAD pre-training followed by tuning on pilot dataset.

***Vanilla.*** We train the models with Adam optimiser and early stopping on validation data loss, the learning rate of 0.001 and patience of 10 epochs. Given the limited amount of data, we adopt Leave-one-out (LOO) approach to model evaluation. In this setting, the model is always tested on data from one subject and trained and validated on data from remaining subjects. We use 8 subjects for training and 2 for validation. The training examples are extracted using a varying window size of length 10, 30 or 60 seconds (depending on experimental conditions) and step size of 8 samples in cognitive load and 4 samples in baseline condition to correct for dataset imbalance. The validation and test datasets have the same window size as the training set but validation data were extracted with a step size of 32 samples (half a second) for both cognitive load and baseline and the test was extracted with a step size of 64 samples corresponding to 1 second. We use weighted F1-score as a performance metric.

***WESAD Pretraining:*** To estimate to what extent 1D CNN model pre-training can facilitate cognitive load detection we utilise publicly available WESAD dataset[16]. The dataset captures signals from 15 volunteers in stressed, amused and baseline conditions. We selected this dataset because it has been also gathered with the use of Empatica4 device and therefore the required information transfer can be limited to population and task, not the device. We trained the model on 13 subjects from WESAD dataset and validated on 1 and test on 1 subject. The model was trained to classify stressed vs non-stressed (baseline or amused) physiological state. In this two-class problem, the step size was 18 samples for non-stress condition and 12 samples for stress condition. This yields for example around 117000 training examples when the window size is 60 seconds. The signal window size corresponded to the widow size used for cognitive load classification. The pre-training was performed with a default learning rate of Adam optimiser (0.001) in Keras. During fine-tuning for cognitive load detection the models were trained with a lowered learning rate of 0.0001.

*Application Development.* We developed two versions of the mobile survey application, both were written in Kotlin and utilized Jetpack Compose package. The survey application consisted of four surveys: GRIT, Positive and Negative Affect Schedule (PANAS), HAPPINES and Need for Recovery (NFR) relating to mental well-being in a work context. In the standard survey application, participants were shown one question at a time (see Figure 2a) and after the completion of all questions from one survey, they proceeded to the next questionnaire. In gamified survey condition, we added a progress-tracking element in the form of a personalized avatar that walks towards a present (see Figure 2b). Our previous research showed that the majority of participants in our study fall into the *Player* gamer archetype, therefore, we selected game elements associated with this role (progress tracking, points, rewards and levels). The game task is to complete all four surveys. Each filled survey is associated with leveling up. After completion of each level, the participants see a congratulation screen in which they receive a present and points as a reward (see Figure 2c). Game elements personalization was only achieved through tailoring of the avatar.

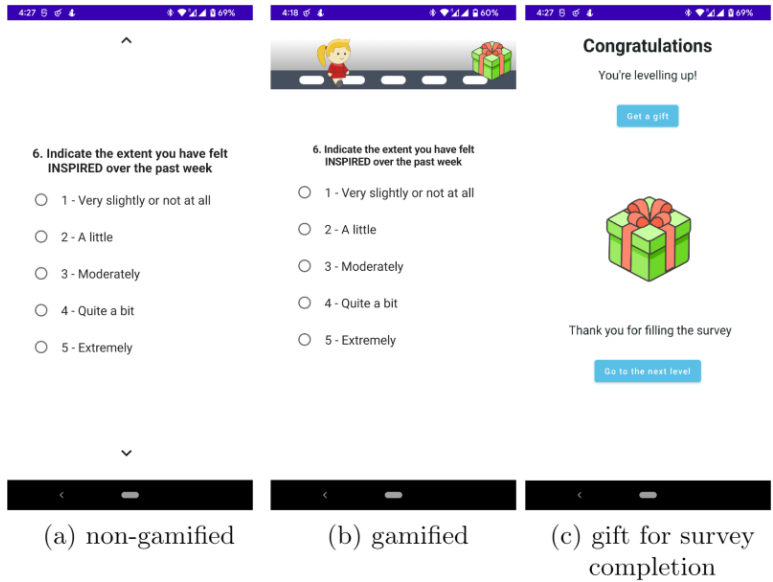

(a) non-gamified  (b) gamified  (c) gift for survey completion

**Figure 2.** Screenshots of question screens in the survey application and a surprise gift acquired for the completion of every survey in the gamified version.

## Results and Discussion

**Table 1.** Cognitive load detection performance (mean weighted F1 score and std from 40 training runs) using different training protocols and length of the signal window.

| | Vanilla | | | | | | WESAD Pre-Trained | | | | | |
|---|---|---|---|---|---|---|---|---|---|---|---|---|
| | 10 | | 30 | | 60 | | 10 | | 30 | | 60 | |
| Subj. | Mean | Std | Mean | Std | Mean | Std | Mean | Std | Mean | Std | Mean | Std |
| 0 | 0.745 | 0.070 | 0.669 | 0.301 | 0.703 | 0.238 | 0.880 | 0.014 | **0.902** | 0.018 | 0.873 | 0.133 |
| 1 | 0.564 | 0.119 | 0.478 | 0.291 | 0.483 | 0.290 | 0.587 | 0.122 | **0.680** | 0.163 | 0.624 | 0.323 |
| 2 | 0.508 | 0.094 | 0.471 | 0.109 | 0.512 | 0.030 | 0.464 | 0.002 | 0.483 | 0.015 | **0.564** | 0.115 |
| 3 | 0.544 | 0.098 | 0.595 | 0.236 | 0.656 | 0.178 | 0.610 | 0.108 | 0.692 | 0.189 | **0.833** | 0.172 |
| 4 | 0.361 | 0.068 | 0.259 | 0.070 | 0.329 | 0.099 | 0.530 | 0.101 | **0.570** | 0.128 | 0.524 | 0.109 |
| 5 | 0.346 | 0.123 | 0.260 | 0.216 | 0.391 | 0.286 | 0.572 | 0.107 | **0.617** | 0.148 | 0.591 | 0.204 |
| 6 | 0.901 | 0.038 | 0.685 | 0.341 | 0.794 | 0.248 | 0.903 | 0.038 | **0.916** | 0.074 | 0.840 | 0.184 |
| 7 | 0.411 | 0.129 | 0.346 | 0.190 | **0.436** | 0.215 | 0.378 | 0.185 | 0.222 | 0.165 | 0.205 | 0.206 |
| 8 | 0.271 | 0.055 | 0.309 | 0.207 | 0.442 | 0.266 | 0.366 | 0.143 | 0.472 | 0.208 | **0.559** | 0.298 |
| 9 | 0.386 | 0.080 | 0.373 | 0.162 | 0.461 | 0.202 | 0.630 | 0.107 | 0.779 | 0.177 | **0.795** | 0.178 |
| 10 | 0.398 | 0.060 | 0.305 | 0.155 | 0.429 | 0.221 | **0.448** | 0.100 | 0.445 | 0.146 | 0.433 | 0.143 |
| Mean | 0.495 | | 0.432 | | 0.512 | | 0.579 | | 0.616 | | 0.622 | |

*Cognitive Load Detection.* To understand how to obtain a reliable cognitive load detection model we investigate different approaches to training protocols including varying lengths of the signal input and model pretraining as described above. Due to a small number of data, the model performance was varying between runs and therefore we run each training session 40 times and report the mean and standard deviation of all runs. We find that the number of data samples was insufficient to train the model from scratch on the pilot cognitive load dataset only (see **Vanilla** in Table 1). Model pre-training on WESAD dataset has shown to be helpful for cognitive load detection (see **WESAD pre-trained** in Table 1). The length of the signal is also important, in general, the 10 seconds window seems to be insufficient to obtain promising cognitive load detection performance even when the model is pretrained.

We expected that the better the performance of the stress detector on the pretraining task the better will be the performance on the fine-tuning task. However, the link between source and target task performance is not that apparent. There is a weak correlation between source training data stress detection and fine-tuned cognitive load detector performance (r=0.18 p=0.04), a slightly stronger correlation between source validation data stress detection and target cognitive load detection (r=0.25, p=0.005), and no correlation between the test subject stress detection and target dataset cognitive load score (see Figure 3). Further, to check if our cognitive load detector is not just detecting stress, we run the best-performing stress detector with a 30s window on all participants from the pilot dataset and check % of samples for each participant from each condition (baseline and cognitive load) that are classified as stressed. We find that signal from participants 1, 3, 4, 7 and 8 had 0 samples classified as stressed, participants 0, 6, 9 and 10 had 9-22% samples classified as stressed and only participants 2 and 5 showed higher levels of stress with 59% and 55% of signal classified as stressed respectively. This suggests that improvements obtained via pre-training are not due to the physiological signal similarity in both conditions (cognitive load is not classified as stress) but rather due to learning of general feature extraction from the 1D signal.

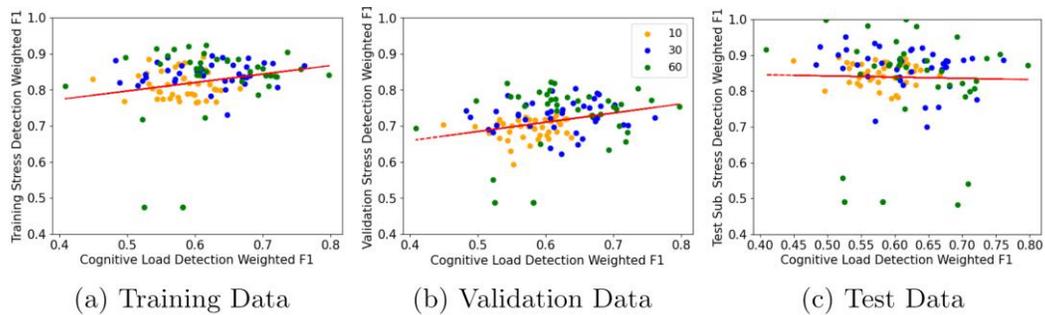

**Figure 3.** WESAD pre-training model performance vs target task performance. The colors represent the signal window size length.

*Response Time.* On average participants spent less time responding to questions in gamified condition than in non-gamified condition, with 5.5s and 6s per question respectively. However, interestingly, in gamified condition participants spent more time at the start and end questions which potentially could be due to the introduction of the gamified components which could capture respondents' attention (see Figure 4). Note that time spent on the very last question in each survey has been not included in this analysis as they mostly capture the time needed to move between questionnaires.

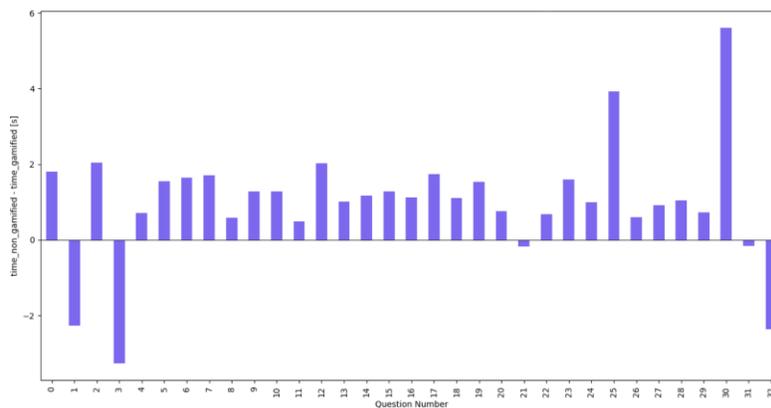

**Figure 4.** Difference between time spent responding to questions non-gamified and gamified survey application.

**Table 2.** Selected cognitive load detector performance and stress level during calibration tasks and percentage of time spent in high cognitive load and stress during filling the gamified and non-gamified surveys.

| Subj. | Session 1 | | | | | Session 2 | | | | |
|---|---|---|---|---|---|---|---|---|---|---|
| | Calibration Tasks | | Survey | | | Calibration Tasks | | Survey | | |
| | CogLoad Detector F1 | Stress % | Gamified | CogLoad % | Stress % | CogLoad Detector F1 | Stress % | Gamified | CogLoad % | Stress % |
| 11 | 0.47 | 0 | yes | X | 6 | 1.00 | 0 | no | 15 | 0 |
| 12 | 0.85 | 54 | yes | 83 | 19 | 0.94 | 10 | no | 100 | 4 |
| 13 | 1.00 | 10 | no | 89 | 1 | 1.00 | 1 | yes | 95 | 7 |
| 14 | 1.00 | 9 | yes | 84 | 2 | 0.77 | 0 | no | 31 | 0 |
| 15 | 0.93 | 0 | yes | 34 | 0 | 0.82 | 0 | no | 100 | 0 |
| 16 | 0.77 | 0 | no | 66 | 17 | 0.71 | 0 | yes | 77 | 7 |
| 17 | 1.00 | 1 | no | 100 | 0 | 0.99 | 0 | yes | 100 | 0 |
| 18 | 1.00 | 66 | no | 100 | 24 | 1.00 | 60 | yes | 95 | 50 |
| 20 | 0.49 | 0 | no | X | 0 | NA | | | | |
| 21 | 0.65 | 17 | no | X | 52 | NA | | | | |
| 22 | 0.99 | 1 | no | 58 | 1 | 0.89 | 2 | yes | 99 | 0 |
| 23 | 1.00 | 18 | no | 94 | 1 | 0.86 | 0 | yes | 94 | 0 |
| 24 | 0.89 | 80 | yes | 47 | 48 | 1.00 | 63 | no | 9 | 41 |

*Cognitive Burden.* To ensure that the cognitive load detector can distinguish between high and low cognitive load for each participant in the survey dataset we first applied all 440 models (from 40 runs and 11 folds) trained with a 30s window on the pilot dataset to every participant in the survey dataset in cognitive load and baseline conditions and for each participant, we selected a model showing the best performance for this participant. In other words, we matched the models to participants using a small calibration dataset obtained from participants prior to filling the survey application. The performance of the selected best model is reported in Table 2 for each session. We have excluded the prediction of cognitive load during survey completion for those participants for whom we could not get a model with performance above 0.7 F1 score (marked with X in Table 2). Encouragingly, for 10 out of 13 participants, we were able to find a model which could reliably distinguish between their high and low cognitive load. The time spent in the high cognitive load and stress does not differ between gamified and non-gamified survey conditions. Potentially to understand if there is a difference between cognitive burden we would need a classifier trained on at least three different tasks with cognitive load levels: high (demanding task), medium (e.g. reading task) and low.

*Participants Experience.* Only half of the participants noticed the difference in survey application appearance between the sessions. Participants who noted the difference reported preferring the gamified version more. Potentially the gamification was too simplistic to modulate the level of engagement or enjoyment. It is nontrivial to introduce game components in the surveys to reduce the burden but not affect the responses.

**Limitations**

The main limitation of this work is the small sample size of participants used for both developing the cognitive load detector and evaluating the cognitive effort required to complete the two survey versions. To mitigate this issue, we pre-trained the detector on a similar task, but further improvements in its performance may be possible with a larger sample size. The generalizability of our method can also be impacted by the homogeneous nature of the study participants. The volunteers were healthy individuals from a similar age group. To ensure the robustness of our method, future studies should include a more diverse and bigger sample size. Finally, we conducted experiments in controlled laboratory settings. In the future, we plan to conduct similar studies in real-world conditions[7].

In this feasibility study, we focused solely on developing automatic detection of high cognitive load, which is the first step towards estimating cognitive effort in the real world. This may not be sufficient to distinguish more subtle differences in cognitive demand caused by the mobile application. To identify the game element or combination of such elements that is the most helpful in reducing cognitive burden we would like to conduct an ablation study in the context of well-being questionnaires in our future work. We would also like to include the self-reported cognitive load after survey completion for a better understanding of survey difficulty. Moreover, it would be interesting to collect data during semi-demanding tasks, such as text reading, in order to train a more finely-graded classifier. Finally, we would like to include other tasks requiring a high cognitive load to ensure the efficiency of our method across various cognitively demanding activities.

**Conclusion**

We developed a cognitive load detector and two versions of mobile surveys to investigate if gamification can reduce the burden of self-reporting. We have found no impact of the addition of simple game elements such as: progress tracking, avatar, rewards on the amount of time spent in high cognitive load or stress during filling in the surveys. However, this feasibility study yields practical learning related to cognitive load model training, such as: 1) Performance of CNN-based cognitive load detector from PPG signal is boosted via transfer learning on stress detection task. 2) There is a link between the model performance on the source and target task. 3) The minimum length of signal for cognitive load classification is 30 seconds but the addition of extra temporal context can further boost the detection. 4) Matching models to the participants using a small calibration dataset can facilitate finding a detector that can reliably distinguish between high and low cognitive load for each individual.

**Acknowledgment**

This work is supported by the European Union's Horizon 2020 research and innovation programme under grant agreement Sano No 857533 and the International Research Agendas programme of the Foundation for Polish Science, co-financed by the European Union under the European Regional Development Fund.

**References**

1. Adamczyk, P., Kuś, M., Marek, S., Pręcikowski, R., Grzeszczyk, M., Malawski, M., Lisowska, A.: Designing personalised gamification of mhealth survey applications pp. 224–231 (2023). https://doi.org/10.5220/0011603800003414
2. Carlier, S., et al. : Investigating the influence of personalised gamification on mobile survey user experience. Sustainability 13(18), 10434 (2021)
3. Chow, C.Y., Riantiningtyas, R.R., Kanstrup, M.B., Papavasileiou, M., Liem, G.D., Olsen, A.: Can games change children's eating behaviour? a review of gamification and serious games. Food Quality and Preference 80, 103823 (2020)
4. Gan, D.Z., McGillivray, L., Han, J., Christensen, H., Torok, M.: Effect of engagement with digital interventions on mental health outcomes: a systematic review and meta-analysis. Frontiers in digital health 3 (2021)
5. Goldberger, A., Amaral, L., Glass, L., Hausdorff, J., Ivanov, P. C., Mark, R., ... & Stanley, H. E. (2000). PhysioBank, PhysioToolkit, and PhysioNet: Components of a new research resource for complex physiologic signals. Circulation [Online]. 101 (23), pp. e215–e220
6. Grzeszczyk, M. K., Blanco, R., Adamczyk, P., Kus, M., Marek, S., Pręcikowski, R., & Lisowska, A. (2023). CogWear: Can we detect cognitive effort with consumer-grade wearables? (version 1.0.0). PhysioNet
7. Grzeszczyk, M. K., Lisowska, A., Sitek, A., Lisowska A.: Decoding Emotional Valence from Wearables: Can Our Data Reveal Our True Feelings?. In 25th International Conference on Mobile Human-Computer Interaction (MobileHCI '23 Companion), (2023)
8. Gwizdka, J.: Using stroop task to assess cognitive load. In: Proceedings of the 28th Annual European Conference on Cognitive Ergonomics. pp. 219–222 (2010)
9. Johnson, D., Deterding, S., Kuhn, K.A., Staneva, A., Stoyanov, S., Hides, L.: Gamification for health and wellbeing: A systematic review of the literature. Internet interventions 6, 89–106 (2016)
10. Lisowska, A., Wilk, S., Peleg, M.: Catching patient's attention at the right time to help them undergo behavioural change: Stress classification experiment from blood volume pulse. In: International Conference on Artificial Intelligence in Medicine. pp. 72–82. Springer (2021)
11. Lisowska, A., Wilk, S., Peleg, M.: Is it a good time to survey you? cognitive load classification from blood volume pulse. In: 2021 IEEE 34th International Symposium on Computer-Based Medical Systems (CBMS). pp. 137–141. IEEE (2021)
12. Markova, V., Ganchev, T., Kalinkov, K.: Clas: A database for cognitive load, affect and stress recognition. In: 2019 International Conference on Biomedical Innovations and Applications (BIA). pp. 1–4. IEEE (2019)
13. Oakley-Girvan, I., et al. : What works best to engage participants in mobile app interventions and e-health: A scoping review. Telemedicine and e-Health (2021)
14. Rim, B., Sung, N.J., Min, S., Hong, M.: Deep learning in physiological signal data: A survey. Sensors 20(4), 969 (2020)
15. Saganowski, S., et al.: Consumer wearables and affective computing for wellbeing support. In: MobiQuitous 2020-17th EAI International Conference on Mobile and Ubiquitous Systems: Computing, Networking and Services. pp. 482–487 (2020)

16. Schmidt, P., Reiss, A., Duerichen, R., Marberger, C., Van Laerhoven, K.: Introducing wesad, a multimodal dataset for wearable stress and affect detection. In: Proceedings of the 20th ACM International Conference on Multimodal Interaction. pp. 400–408 (2018)
17. Stoet, G.: Psytoolkit: A software package for programming psychological experiments using linux. Behavior research methods 42(4), 1096–1104 (2010)
18. Stoet, G.: Psytoolkit: A novel web-based method for running online questionnaires and reaction-time experiments. Teaching of Psychology 44(1), 24–31 (2017)
19. Xu, L., Shi, H., Shen, M., Ni, Y., Zhang, X., Pang, Y., Yu, T., Lian, X., Yu, T., Yang, X., et al.: The effects of mhealth-based gamification interventions on participation in physical activity: Systematic review. JMIR mHealth and uHealth 10(2), e27794 (2022)
20. Zanelli, S., El Yacoubi, M.A., Hallab, M., Ammi, M.: Transfer learning of cnn-based signal quality assessment from clinical to non-clinical ppg signals. In: 2021 43rd Annual International Conference of the IEEE Engineering in Medicine & Biology Society (EMBC). pp. 902–905. IEEE (2021)